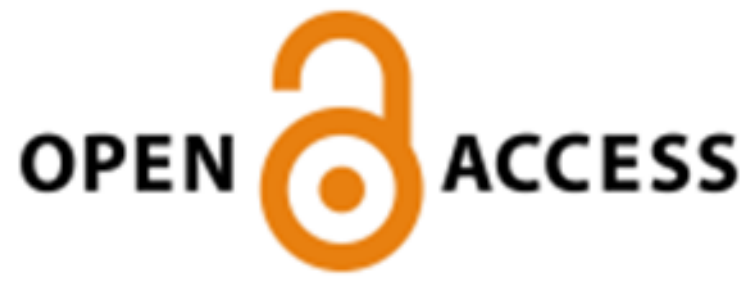

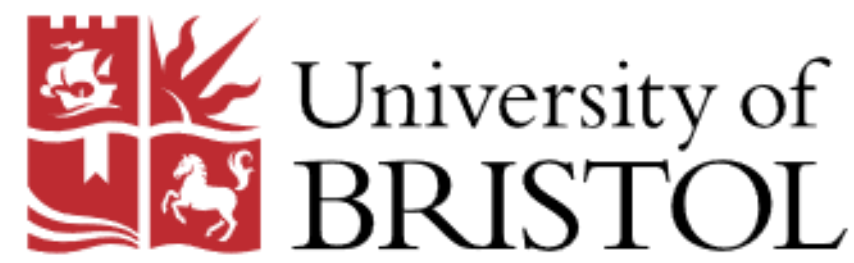

# Active Trust Management for Successful Human-Robot Teaming: Moving from a Trust Repair to a Trust Satisficing Perspective

**Nicola Webb & Edmund R. Hunt**

School of Engineering Mathematics & Technology, University of Bristol

Corresponding Author Email address: edmund.hunt@bristol.ac.uk

**ABSTRACT:**

Integrating mobile robots into human teams promises significant capability improvements for tasks such as searching hazardous environments. Unlike existing teleoperated robots, future robot systems will increasingly be endowed with some level of artificial intelligence (AI), giving them a degree of autonomy in how they pursue mission goals. This autonomy could make a human-agent (robot) team more effective but also put inter-agent trust under strain if robots make a mistake, or (appear to) pursue task priorities that conflict with the team's best interest. During a mission, agents' trust states are anticipated to vary according to the situation as understood by each teammate (trustor). If component-level (agent) or system-level trust falls below sufficient levels for cooperative tasks to be completed, it could critically affect mission success . We argue that active *trust management* will be an important precondition for the success of human-robot teams (HRTs, a subcategory of human-agent teams with embodied agents), especially in dynamic, high-risk environments. We present a trust satisficing perspective which acknowledges and attempts to account for the fluctuating, multi-faceted, and context-dependent nature of trust and trust requirements even under normal operating conditions. Our outline of a trust management framework for human-robot teaming includes online measurement of proxy metrics for trust, closed-loop adaptation of robot behavior, and variable autonomy to give space for human responsibility in situations requiring value judgements. We refer to a recent experimental exploration of 'swift trust' and a novel behavioral trust metric for HRT, and we highlight issues for further investigation.

**INTRODUCTION:**

Robot technologies are recognized as having important potential for use in high-risk environments, for instance in emergency services work, such as search and rescue or post-disaster survey scenarios. This is due to the challenging conditions facing workers in such situations and the possibility of reducing personal safety risks while enhancing effectiveness (Delmerico et al., 2019). In the future, robots are anticipated to have a certain level of autonomy – beyond teleoperation – as they work alongside humans in human-robot teams (HRTs). We focus on HRTs as a distinct subcategory of human-agent teams (HATs), where the agents are physically embodied in the mission environment and can interact with humans through non-verbal communication (e.g., dynamic use of space through time; joint attention through gaze). The question of how to achieve high HAT (HRT) performance is an ongoing and growing subject of research, and establishment and maintenance of trust is regarded as a central element of the answer (e.g., Chen et al., 2014; De Visser et al., 2020; Huang et al., 2021). In this chapter we outline elements of an active trust management framework in HRT, whereby trust can be measured and managed online during a mission so that the mission can have its best chance of success. First, we briefly describe our approach to trust in HRTs, the sorts of high-stakes environments we are focused on with deployment of mobile robots, and the way such environments put trust under strain even in the absence of obvious event-based ruptures in inter-agent trust. While much HRT research has focused on trust repair (e.g., Baker et al., 2018; Kox et al., 2021), we see this research as one component of the broader trust management challenge.

Parts of this chapter have been previously published in the following CC-BY article:

Milivojevic, S., Sobhani, M., Webb, N., Madin, Z., Ward, J., Yusuf, S., Baber, C., & Hunt, E. R. (2024). Swift Trust in Mobile Ad Hoc Human-Robot Teams. In *Proceedings of the Second International Symposium on Trustworthy Autonomous Systems* (pp. 1-10), Austin TX, United States, September 17–18, 2024.

We acknowledge support from the £1.2M UK EPSRC grant “Satisficing Trust in Human-Robot Teams” (2023-26, reference EP/X028569/1).

**Dynamic trust and trust requirements**

A recent review by Malle & Ullman (2021) suggests that trust is multidimensional, incorporating both performance aspects (which are central in the human-automation literature) and moral aspects (which are more often studied in the human-human trust literature). For our research project, “Satisficing Trust in Human-Robot Teams” (funded by UK EPSRC, 2023-26), we have taken the model of Lewis and Marsh (2022), which considers trust to comprise three main components: capability, predictability, and integrity. Integrity effectively covers the moral component identified by Malle & Ullman (2021). As recently noted by Schroepfer and Pradalier (2023), trust is a challenging concept to define and has therefore spawned a diversity of definitions reflecting the disciplinary diversity of human-robot interaction (HRI) research. Broadly, though, trust tends to be conceptualized either with respect to the internal trust state of the trustor (a ‘psychological’ approach) or with respect to situated behaviors and interaction with various external factors (a ‘sociological’ approach). The approach we propose here focuses more on the latter approach, attempting to define 'trust' in a way that is interpretable, and hence usable, by both humans and robots (Hunt et al., 2023), though it includes scope for psychophysical proxy trust metrics (Figure 1, 2). For a practical operationalization of trust concepts, one might identify a set of factors that are associated with high, or at least sufficient, trust between individuals in an HRT (Hancock et al., 2020), ideally in a format that is meaningful to both human and robot team members (Hunt et al., 2023). Notwithstanding pre-existing trust relationships between members of an HRT, we expect trust to vary according to the developing situation encountered by each teammate (Hunt et al. 2023). Huang et al. (2021) use the term distributed dynamic team trust ('D2T2'), "to include both interpersonal factors and technical factors related to trust in human-AI-robot teaming along a dynamic timeline, all of which are typically missing in traditional dyadic trust research." (Huang et al., p.302). Wildman et al. (chapter XX in this volume) also emphasize the critical importance of better understanding the temporal dynamics of trust. Moreover, beyond the immediate HRT on the mission, there is likely to be a command hierarchy and a variety of other stakeholders (e.g., Huang et al. 2021 Schroepfer et al. 2024). These other human stakeholders may also require trust satisficing actions by the mission participants, and so our framework includes space for these actors as well. In our active trust management framework, we try to consider relevant factors as control inputs to an autonomous system able to estimate and *satisfice* trust levels over the course of a mission and perhaps use them as the basis of explanation of robot decisions for the humans. For instance, in Webb et al. (2024) we examined a possible behavioral metric for trust (inter-agent movement synchronization) that could be estimated online during deployment and used as such a control input. Notably, this behavioral metric is distinctive to embodied agents, as an example of the unique considerations of HRT we focus on here.

We have outlined aspects of our overall approach in Hunt et al. (2023), where we consider categories of metrics for which each member of a team is able to acquire information ('When', 'Where', 'Who', 'What', 'Why'). We enumerate some of the different ways teammates (human or robot) can acquire this information, with a view to modelling trust in a way that is mutually intelligible between humans and robots, as far as possible. Without this transparency, there is greater likelihood of members being viewed as unreliable or misunderstood (Sanders et al. 2014). The metrics are similarly applicable, except for 'Why', where values inform human choices. In Hunt et al. (2023) we question whether autonomous value alignment for robots in high-risk environments is plausible, given that interpretation and expression of values in morally sensitive situations (such as those commonly confronted by emergency or defense workers) is a human responsibility. Thus, we see in our framework that the ongoing responsibility to maintain the value alignment of the robot goals and tasks with the teams' values (construed in its broadest sense, beyond team members 'on the ground' to include command hierarchy, political decision-makers, etc.) lies with the human team members, notwithstanding attempts to somehow program a sense of values into robots before deployment. Apart from trust levels being dynamic, our perspective also highlights that *trust requirements* are dynamic and driven by the demands of the mission environment. Thus, for example, while trust between a human and a robot may be sufficient for the dyad to carry out a task in one set of circumstances, in another context – such as the aforementioned morally sensitive situation – trust may need to be higher before cooperation is possible. Thus, 'satisficing trust' does not mean merely repairing or restoring impaired trust but also boosting it to sufficient levels as the mission requirements and situation evolve in unpredictable ways (Hunt et al., 2023). Moreover, it may also include reducing trust *requirements* to meet the situation as it stands, by using variable autonomy to hand (temporarily) a greater degree of control to human teammates, to augment capabilities with expert operation or to reassign tasks, for instance.

We discuss the importance of a Concept of Operations for variable autonomy in more detail in Section ‘Determine a Concept of Operations (CONOPS) for Variable Autonomy’ and we suggest that determining this Concept of Operations is likely to increase system-level trust in the HRT in advance of deployment. This increase in trust is because it will aid in the creation of shared mental models; enhance predictability; manage expectations and reduce uncertainty; clarify roles and responsibilities; and provide a basis for accountability and learning. Wildman et al. (2024) recently put forward a perspective that trust is dynamic and event-based, suggesting research should explore discontinuous changes in trust in response to events such as agent- and human-enacted trust violation and repair. Indeed, much trust research has focused on the possibility and effectiveness of trust repair after damage to trust has occurred (e.g., Baker et al., 2018). While we agree that discontinuous changes in trust are clearly important, they do not amount to the full picture of situated trust dynamics. As we describe in our outline framework in this chapter, factors like changing proximity, situation awareness, and the ability to direct values-based task and goal choices are also going to dynamically shape trust within the HAT at the component and system level. As such, we see trust change in HRT (HAT) as comprising both discontinuous and continuous elements, which together need to be monitored and responded to by a range of actors for missions to be successful.

**Ad hoc teams and swift trust**

In Milivojevic et al. (2024), HRT is considered in the context of mobile ad hoc teams, where the trust between team members is likely to be provisional. While development of trust in the longer term is rightly seen as an important element to successful HATs in many contexts, decrements in trust could be particularly problematic in teams that have formed on an ad hoc basis (Ribeiro et al., 2021). Provisional trust might occur quite often in emergency scenarios where team members may not have previously worked together, where a certain degree of ‘swift trust’ will need to be assumed to give team members the necessary confidence in each other to cooperate (Meyerson et al., 1996). Much research on HAT trust has been focused on agent dyads (e.g.,, a human trustor and a robot trustee), with relatively little work done on real experiments with multiple, mobile robots. This includes research on a one human, two-robot constellation. Therefore, Milivojevic et al. (2024) sought to understand the development of swift trust in HRT in scenarios such as search by teams of firefighters each operating more than one robot. The findings of that experimental study inform the framework proposed here, such that all elements are not assumed to be available: for instance, psychophysical measurements such as eye tracking, which can be used as a proxy trust measure, obviously cannot be acquired without sensorized human team members; gaze behaviors such as fixation duration, gaze coordinates, pupil dilation and eye openness have been shown to indicate one’s trust towards an autonomous agent (Hulle et al. 2024).

In our approach to trust satisficing – maintaining a ’good enough’ level of trust – we envisage that each team member will form estimates of teammates’ capability, predictability, and integrity as task performance is observed directly or reported by other teammates (Hunt et al., 2023). As robotics continues to mature as a technology, people will encounter an ever-increasing variety of robots, provisioned in larger multi-robot teams, and thus may have little opportunity to build up a long-term trusting relationship with specific individual robots, which foregrounds the importance of the swift trust perspective on HRT. Haring et al. (2021) outline the components of swift trust in an HRT context, which include environmental factors, imported information, individual trust tendencies, and surface-level cues. Imported information refers to pre-existing knowledge that users bring to the ad hoc team, which can include category-based stereotypes associated with the trustee (e.g., stereotypes about their occupation), or from trusted third-party information such as certifications or group memberships (Haring et al., 2021; Wildman et al., 2012). In the case of robots, relevant imported information could include prior interactions with the provider of the robot (e.g., the fire service or military branch) rather than the specific technology at hand.

In the following sections we introduce our trust management framework, commenting on individual elements and giving examples.

**Active Trust Management: An Outline Framework**

Our proposed active trust management framework is shown in Figure 1. We divide the chart into three phases (top): pre-mission (before the HRT is deployed), on-mission, and post-mission. A more detailed view of the on-mission phase is shown in Figure 2. As marked by a red asterisk, some elements are unlikely to be feasible in a 'swift trust' context, i.e., when an ad hoc HRT forms, for example in an emergency when human team members must work with an unfamiliar type of robot.

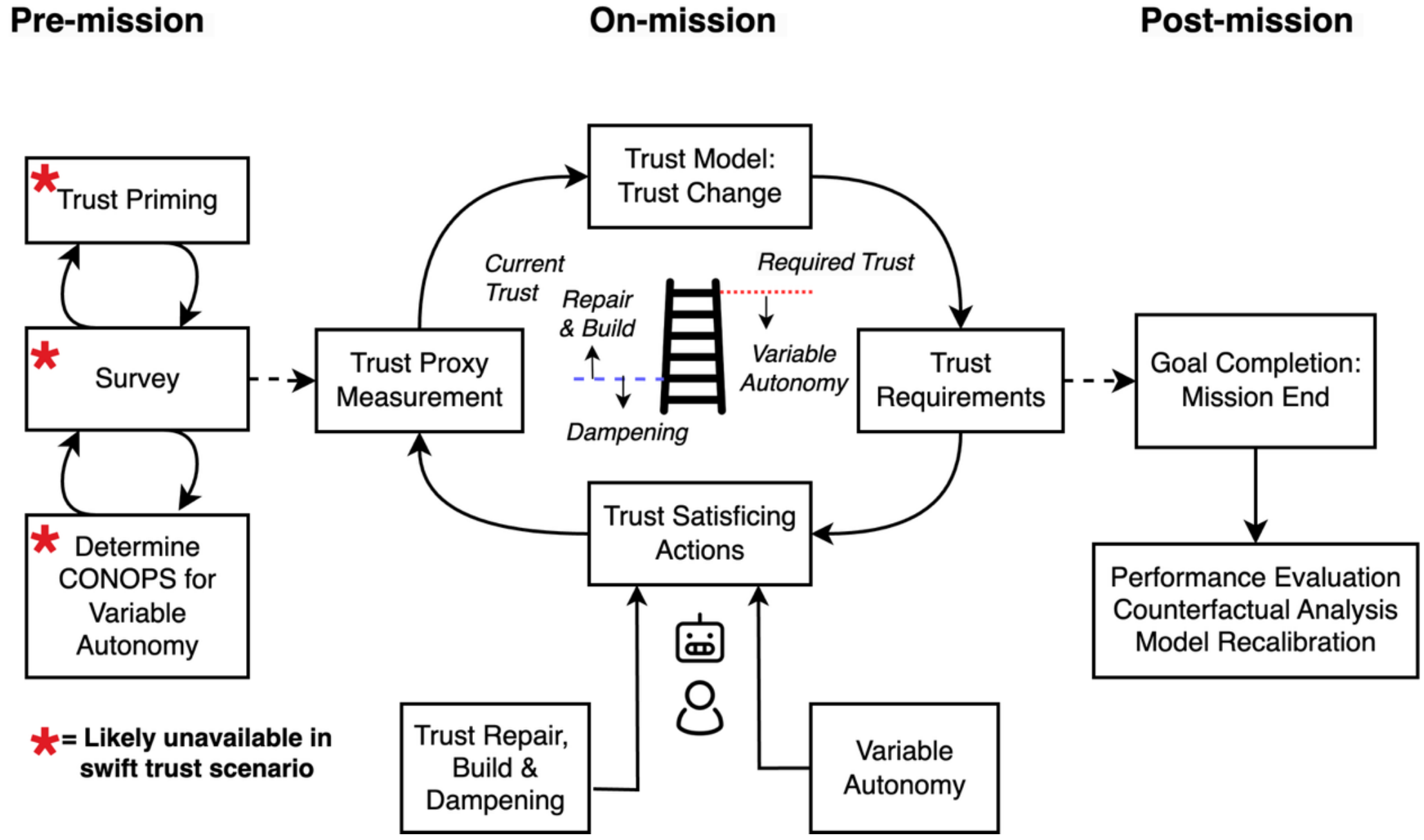

**Fig. 1.** A flow diagram depiction of the stages of our active trust management concept, which begins before deployment. Depending on whether the mission is routine or ad hoc, the 'swift trust' demand will shape the possibility of including all elements. 'Satisficing trust' can involve raising trust levels (at the dyad, team, or system levels) through actions to repair or build it, dampening it prudently if agent capabilities are becoming a poor match for the task, and/or lowering required trust, by taking greater control of the robot through variable autonomy (i.e., lowering the autonomy level).

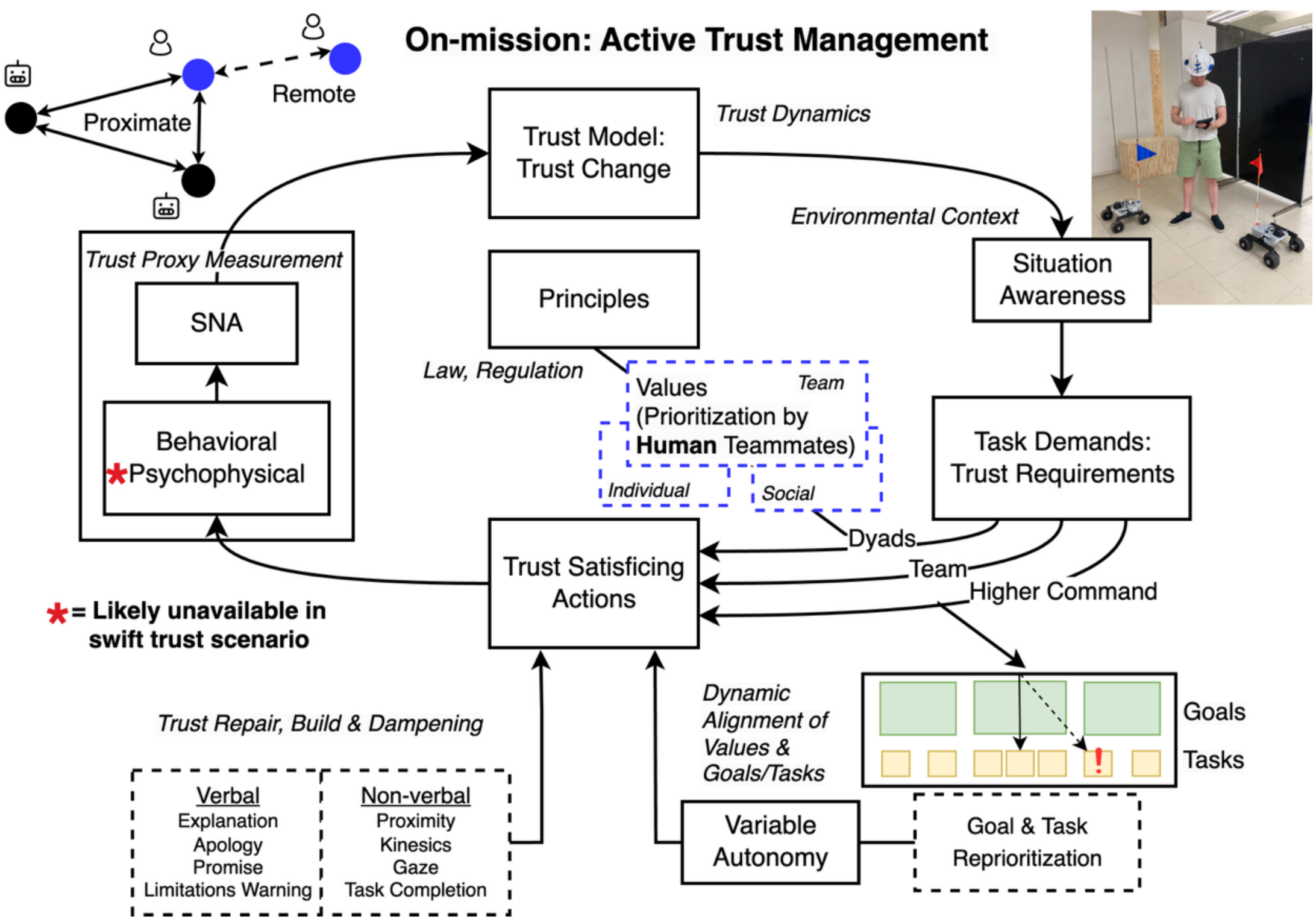

**Fig 2.** This diagram illustrates the central feedback loop of the on-mission phase of active trust management, where trust is continually measured, modelled, and acted upon to meet dynamic mission requirements. An important aspect of HRT trust satisficing is the continued dynamic alignment of the robot and human goals and tasks with humans' individual, team and social values, which requires periodic goal and task reprioritization; this need for dynamic value alignment provides additional motivation for a variable level of autonomy whereby users can intervene with robot goal and task choices where necessary. A non-exhaustive variety of trust repair and build methods are shown.

### *Pre-mission*

In the first phase of the framework, we envisage that human team members can be surveyed for self-reported trust levels in the robot(s) to be used in the mission. Such a survey can be used for trust priming, which essentially focuses on increased transparency as to how the robot or system of robots makes decisions. This transparency is an important aspect of trust calibration (Lee & See, 2004) and ensuring initial over- or under-trust is avoided. The Concept of Operations (CONOPS) for how the HRT will control the robots' level of autonomy for different tasks and situations should also be developed in advance of mission deployment, and it will be informed by the level of trust in the robots and shape trust priming approaches. Depending on how much time is available pre-deployment, these three elements could be iterated to a high level of refinement.

**Trust Survey**. Assuming a mission is routine, the human team members will know in advance which robot platforms are being used. This advanced knowledge means that baseline levels of trust can be established before deployment through (rapid) survey instruments. Self-reported measures, such as questionnaires, are a common method for assessing trust. One notable example is the 14-item Trust

Perception Scale developed by Schaefer in 2016. This scale is designed to capture individuals' subjective trust perceptions across various contexts, providing insights into how trust is established and maintained in human-robot interactions. However, relying solely on self-report measures can be problematic. Chita-Tegmark et al. (2021) consider different trust measures and investigate which items in the questionnaires are suitable for which robots and interactions. They found people have differing mental models of robots, and so one must be mindful not to make assumptions regarding how users will anthropomorphize robots, for instance, which will affect how they will trust them. They also found that people can define words differently and may find some survey questions ambiguous. Such limitations could be partly mitigated through consistent pre-survey briefing and instructions, and definition of terms.

**Trust Priming**. Creating a foundational level of trust before initiating teamwork is essential for fostering effective collaboration among team members. One effective approach to building this trust is known as *trust priming*, which involves taking deliberate steps to shape users' mental models of an AI or robot in a way that increases the likelihood of positive outcomes (Pataranutaporn et al., 2023). Trust priming differs from priming in a more traditional sense, which involves the activation of an existing mental model as opposed to altering them. A key component of trust priming is ensuring that all team members have a clear understanding of how a robot or automated system makes its decisions, often by providing insights into the algorithms, logic, and data that inform the robot's actions through training. Before setting out on a mission, team members can be trained on the agents' abilities and setting expectations about where an agent may fail. The timescale over which this understanding is best achieved (e.g., hours, days, or weeks prior) is an important question for future research. When team members are informed about a robot's decision-making process, they are more likely to feel confident in the robot's capabilities, increasing their trust in both the technology and one another.

Enhanced understanding not only boosts trust but also contributes to improved overall decision-making and performance within the team. When team members are aware of what to expect from the robot, it allows them to align their goals and collaborate more seamlessly. For example, Pataranutaporn et al. (2023) found that priming individuals with information about AI agents (an LLM-based chatbot in their case) can influence mental models and hence shape perceptions of trustworthiness, empathy, and performance, which in an HAT or HRT context could favor more effective teamwork. As Pataranutaporn et al. note, 'sociotechnical imaginaries' can have an important narrative influence in shaping shared mental models about what endeavor the human-AI (robot) team is embarked upon. This term refers to the feedback loop between humans' imagination of the future and present social reality. To attempt a simple example in the HRT search and rescue case, a human user's imagination that a robot is a loyal friend like a trained dog will not correspond to the engineering reality of its simple OODA loops (observe-orient-decide-act). More than that, the human's imagination of his or her place within a technological vanguard, and that future vision's attendant norms and values, will shape behavior in the present situation. Therefore, trust priming could include deliberate actions to understand and recalibrate sociotechnical imaginaries, recognizing the broader social context for how people engage in 'algorithmic reading' (Finn, 2018).

**Determine a Concept of Operations (CONOPS) for Variable Autonomy.** CONOPS for how the HRT will work together on the mission, particularly in complex and ethically charged environments, should specifically address the use of variable autonomy to ensure that robots function in accordance with relevant human values (individual, team, social) and appropriate to the level of trust in them. For instance, when the mission environment creates dilemmas such as continuing to pursue a goal-related task (e.g., surveil a target building) or assist the evacuation of a civilian, an autonomous robot should ideally escalate the decision to a human teammate. The CONOPS would define the specific responsibilities of the human team members, including not only supervision and possible teleoperation, but also the extent of their moral and legal responsibility for the robots' actions. It would define a clear spectrum of autonomy, ranging from full autonomy for well-understood and defined tasks (e.g., search a building), to supervised autonomy or full teleoperation for more complex or morally fraught situations. Critically, it should specify (to some level of precision) the conditions under which the level of autonomy will change, especially if the expectation is that the system itself should trigger this change according to quantifiable thresholds.

Triggers to change autonomy level will interact with the level of trust in the robot actors. For example, in situations with low trust and high moral consequence (e.g., an unreliable drone navigating near civilians), the CONOPS would mandate a shift to a lower level of autonomy – either from proactive human intervention or through the autonomy itself handing back control. In a situation of high trust and low moral consequence (e.g., a slow rover mapping an empty building) the CONOPS might permit full autonomy. The CONOPS would also define how and what information the robots communicate to their human teammates – so they can maintain situation awareness – including when to explicitly signal uncertainty, or when to flag up a morally fraught situation (c.f. Verhagen et al., 2024). Finally, it would contain explicit rules of engagement making clear what robot actions are permissible, interaction protocols (with the robot's environment, other robots, and humans), and operational constraints (physical and logical limits), such that the human teammates understand where their own and others' responsibility lies (immediate users, commanders, technicians, manufacturers, etc.). In the case where humans are working with unfamiliar robots in an ad hoc team, establishing this CONOPS before deployment will be highly constrained, and in that regard some kind of standardization of approach within application domains (e.g., emergency services, defense) may be highly valuable.

***On-mission***

A more detailed view of the on-mission phase of active trust management is shown in Figure 2. In the on-mission phase, we envisage an ongoing loop of trust proxy measurement, trust modelling, awareness of changing trust requirements and, where necessary or recommended, trust satisficing actions by the robot or human team members, either individually or as a concerted (sub)team. We first consider trust metrics that are more focused on the individual or dyad (behavioral and psychophysical) and then the system-level (social network analysis). In our framework, we do not include survey pop-ups as a measure (noted in Huang et al. 2021) because while these could be useful for experimental calibration of unobtrusive trust metrics, they would be unwelcome in a real mission deployment, at least in high-risk contexts.

**Behavioral trust metrics.** Behavioral measures are frequently employed to evaluate trust, drawing from a range of observable indicators. One significant aspect of these measures is team performance assessment (Kohn et al., 2021), which involves analyzing how effectively team members—both human and robotic—collaborate to achieve common goals. Metrics such as the accuracy of task completion, quality of outcomes, and overall team productivity are critical for understanding the functioning of the team as a cohesive unit. Another important aspect of behavioral measures is the evaluation of decision-making and response times. These metrics offer insight into how swiftly decisions are reached and how quickly team members react to changes or challenges during tasks. Shorter decision times in the context of interdependent tasks can reflect higher levels of trust, as team members are likely to feel more confident in each other's abilities. Furthermore, the extent to which team members depend on a robot and the frequency of human interventions during tasks can reveal underlying trust dynamics. A higher reliance on robotic support coupled with fewer interventions may indicate a robust level of trust in the robot's capabilities. In Hunt et al. (2023), we relate these to capability and predictability, per the Lewis and Marsh (2022) model.

Nonverbal or implicit communication are important for increasing the transparency and understandability of a robot's internal state, for improving task performance, and for recovering from errors quickly (Breazeal, 2005). Nonverbal communication is a core aspect of human-human communication (Urakami, 2023) and thus understanding the range of nonverbal cues that might be afforded, by design or incidentally, by robot behavior is an important opportunity for HRT. In particular, proxemics and kinesics are important modalities for human-robot interaction (Saunderson, 2019).

In Webb et al. (2024), we considered what observable behavioral changes in movement characteristics can be identified during an HRT mission, which might be markers of (impaired) implicit communication, and hence be used as indicators of trust development. In an experimental study with 22 participants, we tracked whole-body movement dynamics of team members (one human and two robots) in an extensive test environment of approximately 200 square meters using an ultrasound tracking system. Using pre- and post-mission trust surveys (Schaefer, 2016), we found a correlation between trust change and movement synchronization between the human and the most used of the two robots. We posit that in the experimental condition with the deliberate interruption to explicit communications, there was a (variable amount of) damage to the participants' trust in the robots, especially the one they were using at the time of interruption. One participant was quoted speaking of the loss of communication: "[The robot] didn't come to help us. No. If that was a real gas leak, I'd be dead. … Once you've broken the trust, there is no trust." This loss of trust, in turn, is hypothesized to have impaired participants' capacity for subsequent implicit communication after communications had been restored, specifically coordination, for which movement synchrony is a key indicator. We suggest that synchrony can also build trust, which would favor improved communications. In this view, trust acts as a mediating factor between reliable communication and the ability to achieve behavioral synchrony (a distinct construct in social psychology, communications studies, and elsewhere). Qualitative interviews with participants indicated that robot proximity was an important factor in which robot they preferred to use and that a robot staying nearby over time made it "part of the team." On the other hand, when communications were interrupted, the robot being nearby and yet remaining unresponsive aggravated participants' feelings of frustration (Webb et al., 2024). Beyond the opportunity for such a behavioral trust metric to indicate the need for trust repair, we also propose that it could be used as an 'early warning signal' for when trust levels might be slipping, but not in an obvious event-based way (e.g., an identifiable mistake by a robot).

Cha et al. (2018) surveyed nonverbal signaling methods for non-humanoid robots, which are the type nearly always deployed in realistic security, defense and emergency scenarios. Their goal was to “support varying levels of interaction while increasing transparency of a robot’s internal state” (Cha et al., 2018, p. 214). They identified different levels of interaction, starting with mere ‘coexistence’ (where humans are bystanders or observers), rising to ‘coordination’ (coordinating actions in time or space, for greater efficiency or conflict prevention), and then describing the highest level of interaction as ‘collaboration’ (actively working toward a shared goal by communicating such that actions are complementary). In our study (Webb et al., 2024), involving two rovers that purposefully navigate around the perimeter of the experimental space until the user calls them to help, we identified all three such levels of interaction. Users could watch the robots as they ‘autonomously’ search the room (coexistence), walk alongside or nearby a robot with mutual care between the human and robot not to collide with each other (coordination), and also stand alongside a robot for a simulated joint search task (collaboration). We suggest that each of these three increasing levels of interaction are likely to have an increasing impact on trust levels, for better or worse, depending on if the interaction is successful; and that there is likely to be a relationship between decreasing social distance and these higher levels of interaction occurring. We propose that there is significant opportunity for HRT research to identify related indicators of effective ongoing nonverbal communication and use them for active trust management as we outline here. In our future experiments, we will also consider alternative, non-intrusive measures of trust instead of self-reported trust: for instance, by asking the participants to explicitly choose to delegate a sub-task to the robot or to perform it themselves, thus using robot reliance as a proxy trust measure. With respect to robot design, the availability of other forms of minimal, nonverbal communication suited to non-humanoid robots (Cha, 2018), such as indicator lights or sounds, could also be used to increase the legibility of the robots' state and mitigate the trust damage that can occur in the absence of explanation when human-robot teaming dynamics are disrupted.

**Psychophysical trust metrics.** Psychophysical measures of trust provide a physiological perspective on trust, typically using invasive physical measures, which are unlikely to be available in ad hoc teams (where ‘swift trust’ will need to be established). Ajenaghughrure et al. (2020) provide a review of various approaches. Electroencephalograms (EEG) are used to monitor electrical activity in the brain, offering insights into cognitive processes that underpin trust. Certain EEG channel activations have been found to correspond to events in a trustworthiness scenario between humans and agents (Wang et al., 2018). Electrocardiograms (ECG) measure the electrical signals of the heart, which can be indicative of emotional responses linked to trust or mistrust. Waytz et al. (2014) propose that the higher the trust in an agent, the lower the heart rate increase and startle response when faced with an accident involving the agent. Electrodermal activity (EDA) is another valuable psychophysical measure, as it assesses changes in the sympathetic nervous system related to sweat gland activity. This measure can indicate levels of arousal or anxiety during interactions, providing a deeper understanding of the emotional state of team members. Additionally, eye-tracking technology serves as an effective tool to analyzes participants' focal points during interactions, revealing how attention is directed and whether it shifts in response to different stimuli. Patterns in gaze behavior can be telling of trust or mistrust, as individuals may exhibit specific visual cues, such as increased fixation duration, when feeling secure or uncertain (Hergeth et al., 2016).

**Social Network Analysis based trust metrics.** Social network analysis (SNA) can be used to capture trust networks between team members, commanders, and other stakeholders (Huang et al. 2021), which in principle can enable network-based metrics for emergent, system-wide trust (Walliser et al., 2023). It could also be used to estimate trust attitudes of focal individuals. One approach to trust evaluation is described by Gong et al. (2020) based on linear uncertainty theory. Their model allows for computation of the 'recommended trust' between agents in a complex network, whereby the transitivity of trust (if A trusts B, and B trusts C, then A is likely to trust C) points toward trust between agents that do not have direct interaction experience. Such an approach permits estimation of trust relationships between the wide range of stakeholders present in a real-world HRT/HAT deployment, and accounts for incomplete information using uncertainty modelling. Trust estimates from the various agents' trust models (next subsection) can thus be input into such a network analysis, to produce a range of proxy measures for e.g. team and system-level trust that are fed back as evidence into future individual trust model estimates in an iterative manner. For instance, the 'recommended trust' value calculated by the SNA for a pair of agents (A, C) could become an evidence node in agent A's BBN for its trust in C, supplementing its own direct experience.

**Trust modelling and trust change detection**. Using time series of trust measurements, trust change can be monitored via a dual system of trust models for each robot (agent). This dual system distinguishes between a robot's own trust *in others* (a 'trust-out' model) and its inference of others' trust *in itself* (a 'trust-in' model). These models can be maintained in either a distributed (across agents) or centralized manner (e.g., an off-site server), depending on factors like available communications. We have previously considered (Hunt et al., 2023) how trust levels (which are hidden or 'latent' variables) could be inferred using a Bayesian Belief Network (BBN): this methodology has previously shown its utility for modelling distributed situation awareness (Yusuf & Baber, 2022). Under this dual trust modelling framework, each robot (agent) would maintain two BBN models: a 'trust-out' BBN producing estimates of its trust in all other relevant agents, and a separate 'trust-in' BBN for inferring the trust those agents (humans, other robots and AI agents) hold in it. These distinct models are necessary as each direction of trust is inferred from different tangible evidence. Trust estimates would be shared locally to nearby agents or directly to a central command center that would have oversight of the trust levels between agents over time. In a BBN model for trust, the system of HRT members, command hierarchy, and even other stakeholders such as engineers and managers (e.g., Ho et al. 2017), could be modelled using a graph. The BBN graph $G(N, E)$ comprises $N$, a set of nodes connected by causal directional edges, $E$. Each node represents an element of component-level trust with a defined number of states (e.g., goal achieved or not achieved, agent location in different zones, agent reputation positive or negative). Trust proxy measures (e.g. psychophysical, behavioral, social network analysis-based measures, c.f. Figure 2) would be included in the BBN model after thresholding continuous values into discretized states. To provide an example with an SNA measure, the average degree of the communication network over a window of time could provide evidence for the latent (hidden) variable of team cohesion, which one might posit is related to team-level trust. For a given team size, e.g., a five-agent team, the average degree could be thresholded into states 'dense', 'nominal', and 'sparse', with 'sparse' being defined as less than 1.5 for a five-agent team (for instance). Similarly, a behavioral metric like human-robot movement synchrony could be thresholded into states such as 'synchronized', 'nominal', and 'asynchronous', providing another observable evidence node for the latent trust variable. States have associated probabilities, i.e., between 0%-100%, to reflect the assumption that each agent is estimating the chance that the respective states have materialized (this allows for noisy or partial observations). Probabilities are then updated based on mission information or learned over time (Yusuf & Baber, 2022). In this way, the trust elements (e.g., goals, tasks, reputation etc.) can be modelled using BBN nodes with assigned probabilities and causal relationships based on the operating mission context (see further discussion in Hunt et al., 2023). The dual trust-out, trust-in perspective significantly enhances autonomous response capabilities. Via a trust model, be it based on a BBN or otherwise, if a fall in trust is detected – at the level of agent dyads, the team or at a higher command or stakeholder level – trust satisficing actions may be prompted or otherwise commanded from an array of strategies. For example, a trust level decrement in a robot's 'trust-in' model (e.g., decreased task delegation by a human teammate; closer human monitoring of the robot; more manual overrides; lower physical proximity indicating discomfort; expressions of verbal frustration), could prompt trust repair behaviors. A decrement in 'trust-out' for a certain teammate (e.g., actions relating to integrity, such as violating applicable regulations) might trigger risk mitigation behaviors (for example), such as enhanced behavioral monitoring.

**Situation awareness.** An important aspect of active trust management will be maintaining sufficient situation awareness, i.e., an understanding of the mission environment and its various elements, and how they are changing over time as the HRT and other third parties interact with it (Ezenyilimba et al. 2023). Limited situation awareness may lead to a loss of trust if a robot teammate acts in a way that is not understood by human teammates (Webb et al., 2025); it will also make allocation of responsibility more challenging (Waterson et al., 2025). Pertinent information will be gathered via the trust measurements mentioned in the prior subsections, including the spatial positions and communications of team members and their physiological state. For instance, the level of risk in the environment might be revealed by heart rate measurements. Other information will also be relevant beyond the immediate trust metrics, such as the presence of other actors identified via reconnaissance (e.g., satellites). Good situation awareness will allow insight into the changing trust requirements of the various members of the extended HRT.

**Task demands and trust requirements.** As the mission continues and the HRT pursues its goals, various tasks and subtasks will need to be completed. To the best of the team's ability, given its level of situation awareness, it will need to assign humans and robots to these tasks, which will require a certain level of trust between them at the component and system level. Here we come to the notion of 'satisficing trust'. Where the relevant trust levels are insufficient, or at risk of becoming insufficient given the estimated trajectory of trust change, trust satisficing actions may be prompted.

**Trust satisficing actions.** Trust satisficing actions may be led from the bottom up (self-organized with the initiative of individual agents, i.e., triggered by the dual trust models of robots or AI agents, or the action of human teammates) or from the top down (instructed through the command hierarchy, e.g., from a remote command center). The choice of trust satisficing action could be prioritized according to empirical studies of their efficacy (e.g., Baker et al., 2018). While trust satisficing actions will typically be enjoined to the robot team members (i.e., to improve trust attitudes of relevant human teammates toward them), it is possible that human team members may need to take some action to improve the trust of their robot teammates in them as well, depending on their actions and the dynamic environmental context. We have previously introduced the notion of a 'Ladder of Trust' (Baber et al., 2023), as a metaphor for how allocation of function among team members could be dynamic, with tasks potentially being reassigned according to the level of trust agents have in each other to reach the necessary rung on the ladder. Apart from event-driven trust violation and repair events (e.g., Wildman et al., chapter in this book), we identify that variable autonomy (i.e., the human or robot taking, or handing over, greater decision-making control) is an important part of our trust satisficing perspective. Even with a constant level of trust in robot teammates, a dynamic situational context is highly likely to present variable trust requirements that may exceed what an autonomous agent (robot) can (or should) obtain, thus favoring a switch from full to assisted autonomy, for example.

We consider that an important aspect of trust satisficing includes value-driven goal and task reprioritization, because humans retain a vital responsibility to align robot behavior with values (individual, team, social) in the situated context of the mission. This ongoing requirement for value alignment implies the need for some level of variable autonomy, as otherwise autonomous robots are unlikely to be able to satisfactorily resolve morally fraught situations. The trust requirement for a robot to act autonomously in certain situations will be too high to satisfice, and therefore a human will have to either assume greater supervision or teleoperation or otherwise complete the task themselves. In this regard, Verhagen et al. (2024) have discussed ‘meaningful human control’ and variable autonomy, whereby a robot is able to identify when it is a ‘morally sensitive situation’ and reallocate the decision to a human operator. At the center of the trust management loop (Figure 2) are codified principles – applicable laws and regulations – that impose constraints and obligations on behavior. These principles inform values, which are moral in character, but the expression of values in terms of the choice of goal and acceptable outcomes for the HAT mission is context-specific. Although HAT leaders will deploy missions with goals and associated tasks in mind, sometimes these must be adapted online. As we suggest in Hunt et al. (2023), “values can fill the gap where legal principles fail to provide guidance. For example, acting according to social values would be to defuse a bomb even if that means a loss of human life, where such action saves the lives of many others… Principles command; values recommend; goals direct” (p. 1).

For both robot and human team members, we reflect that a range of ‘verbal’ (text or speech) and nonverbal trust repair actions are available. Verbal strategies are a well-studied method of repairing trust after a violation. A review by Esterwood and Robert (2022) highlights mixed findings in this area. Apologies, for instance, tend to be more effective when delivered closer to the trust violation event, but their effectiveness diminishes with time. Denials and explanations also show varying degrees of success, with explanations being less effective when the trust violation is severe. Wildman et al. (chapter in this volume) provide a summary of some recent research on explanation-based trust repair. Promises are often examined in conjunction with apologies and explanations, but their independent impact remains inconsistent. No repair measure has shown to be superior, however more research is needed, especially in regard to the timing of strategy deployment. Non-verbal behaviors play a critical role in trust repair as well. Proximity and synchronized movement patterns could contribute to rebuilding trust: as noted in our earlier discussion of Webb et al. (2024), moving together through shared spaces could foster a sense of cohesion and companionship in human-robot teams. Gaze behaviors also influence trust dynamics. Zhang et al. (2024) found that increased gaze fixation occurs when trust is violated, leading humans to visually scan their environment more actively in an effort to reassess the robot's reliability. Successful task completion is another key factor: Guo and Yang (2021) found that observing a robot successfully complete tasks over time leads to an increase in trust from human collaborators.

While trust repair, build and maintenance have received most research attention, we also recognize that 'over-trust', leading to over-reliance (Lee and See, 2004) may be an issue in human-robot teams. As such, 'trust dampening' actions can also be available to team members; de Visser et al. (2020) list candidate methods for this, including providing timely warnings about a robot (or indeed, a human's) potential limitations in certain emerging contexts (e.g., when a task is potentially too challenging for its capabilities). In this regard, situation awareness of the dynamic mission environment (previous subsection) is likely to be a decisive factor in initiating trust dampening actions.

Overall, recent research findings collectively underscore the importance of addressing both verbal and non-verbal elements to build, maintain, repair and, where necessary, dampen trust in collaborative human-robot teams. Notwithstanding the effectiveness of trust manipulation strategies, the availability of variable autonomy (by user initiative or system handover) is a vital enabler when trust levels fail to meet evolving trust requirements.

***Post-mission***

In the post-mission phase, depending on the series of actions taken and other factors, the mission will end with some degree of success (or failure). We include trust model recalibration and counterfactual analysis (CFA) of the trust satisficing actions as an important part of the mission debrief. These rely on logged time series of agent actions, environment state, and trust metrics and BBN trust estimates, which may be more complete after the mission if disparate agents had only partial knowledge of each other's data. Time permitting, additional data could be obtained through surveys and interviews of human team members on how their trust levels varied during the mission. Based on each agent's chosen actions (e.g.,, the disuse or overreliance on a robot teammate) the accuracy of the estimated trust relationships can be reviewed with a view to calibrating how the different trust metrics (sensor-based proxies, SNA, etc.) are weighted in a multi-modal, BBN-based trust estimation architecture. CFA considers the effectiveness of the trust satisficing actions taken in bolstering trust to a sufficient level – or reducing trust requirements in the case of variations in autonomy level – in the context of all the other actions taken by the HRT, other stakeholders, and third parties. The CFA simulates alternative courses of action ('what if?' questions) to consider if the action recommendations were the most effective available. While an ad hoc HRT might find it difficult to immediately carry out a post-mission analysis in a fraught situation, it could be done over a longer time horizon, provided mission data are retained. A formal, written Concept of Operations (Section 2.1.3) will also serve as a baseline against which the HRT's performance can be evaluated. For instance, if something goes wrong on the mission, the CONOPS can help diagnose the failure – whether technical fault, a team member operating outside their defined role, or a shortcoming in the CONOPS itself. The use of the CONOPS can be instrumental in transforming the collection of individual agents into a coherent system that can build more resilient, long-term trust.

## Conclusion

Human-robot teams are already being deployed in high-stakes environments for missions such as search and rescue. It is widely recognized that maintaining sufficient trust between members of the extended team, including the humans and robots on the ground, remote commanders, and other stakeholders, is a vital part of ensuring the team's success and robotic technology's ongoing adoption (Robinette et al. 2017). We have presented the outline of an active trust management framework, identifying and describing important elements of that framework with respect to a trust satisficing perspective that recognizes the need for distributed, dynamic trust to be 'good enough' for the goals at hand (Hunt et al., 2023). This approach goes beyond a reactive, trust-repair-based strategy to proactively sense when trust might be slipping or trust requirements may have shifted. As part of this approach, we explicitly include space for aligning robot behavior with human values, by retaining control over robot actions in sensitive or otherwise fluid situations with variable autonomy, allowing prohibitive trust requirements to be overcome by human intervention. A thorough Concept of Operations for this variable autonomy is an important foundation for HRT mission success. Our approach to trust measurement focuses on behavioral and psychophysical proxy metrics, and social network analysis to quantify individual and emergent system-level trust relationships (e.g., Gong et al. 2020), although further research is still required to firm up the evidence base for these (Huang et al. 2021, Webb et al. 2024). We also allow for ad hoc teaming and its need for 'swift trust' (Haring et al. 2021, Milivojevic et al. 2024). Given a longer-term opportunity to prepare an HRT for mission deployment, trust priming and the broader social context of sociotechnical imaginaries can also be tactfully addressed. Our active trust management framework reiterates the opportunity to undertake a more subtle trust calibration (Lee & See, 2004) that forms adaptive shared mental models among team members. We look forward to a future where robots (and artificial intelligence) are successfully integrated into robust and resilient human-robot teams, for widespread social benefit.

**Authors' Biographies:**

Dr. Nicola Webb is a Postdoctoral Research Associate at the University of Bristol, supported by the UK EPSRC grant "Satisficing Trust in Human-Robot Teams" (2023-26, reference EP/X028569/1) . She previously worked as a Research Fellow on the "RoboTIPS: Responsible Robots for the Digital Economy" project. Her PhD research at the University of the West of England focused on measuring group social engagement and improving robots' social situation awareness. During her PhD she was a member of the ECHOS research group at the Bristol Robotics Laboratory ("Embodied Cognition for Human-RObot InteractionS").

Dr. Edmund Hunt is a Senior Lecturer (US Associate Professor) in Robotics at the University of Bristol, and Co-Investigator on the 'Satisficing Trust' project. He holds a research fellowship with the Royal Academy of Engineering (2021-26) focused on multi-robot systems for infrastructure security. His research interests include behavioral bio-inspiration, human-robot interaction, and field robotics. His 'field swarm robotics' lab has various research projects investigating the use of robot swarms for environmental monitoring. He received his PhD in complexity sciences from the University of Bristol in 2016.